\RequirePackage{fix-cm}
\documentclass[smallextended]{svjour3}       
\smartqed  

\usepackage{subcaption}
\usepackage{graphicx}
\usepackage{balance}
\usepackage{amssymb}
\usepackage{epstopdf}
\usepackage{listings}
\usepackage{enumitem}
\usepackage{amsmath}
\usepackage{bigstrut}
\usepackage{multirow}
\usepackage{tabularx}
\usepackage{booktabs}
\usepackage[font=small,skip=0pt]{caption}
\usepackage{apacite}

\begin{document}

\title{Propensity-to-Pay: Machine Learning for Estimating Prediction Uncertainty
}

\titlerunning{Short form of title}        

\author{Md Abul Bashar\textsuperscript{a} \and
        Astin-Walmsley Kieren\textsuperscript{ab} \and 
        Heath Kerina\textsuperscript{c} \and
         Richi Nayak\textsuperscript{a} 
}



\institute{\textsuperscript{a}Queensland University of Technology, Brisbane, Australia\\
\textsuperscript{b}Energy Queensland, Queensland, Australia\\
\textsuperscript{c}Ergon Energy Retail, Queensland, Australia\\
\email{m1.bashar@qut.edu.au}\\
\email{kieren.astin-walmsley@energyq.com.au}\\
\email{kerina.heath@ergon.com.au}\\
\email{r.nayak@qut.edu.au}
}


\date{Received: date / Accepted: date}

\maketitle

\begin{abstract}
Predicting a customer's propensity-to-pay at an early point in the revenue cycle can provide organisations many opportunities to improve the customer experience, reduce hardship and reduce the risk of impaired cash flow and occurrence of bad debt. With the advancements in data science; machine learning techniques can be used to build models to accurately predict a customer's propensity-to-pay. Creating effective machine learning models without access to large and detailed datasets presents some significant challenges. This paper presents a case-study, conducted on a dataset from an energy organisation, to explore the uncertainty around the creation of machine learning models that are able to predict residential customers entering financial hardship which then reduces their ability to pay energy bills. Incorrect predictions can result in inefficient resource allocation and vulnerable customers not being proactively identified. This study investigates machine learning models' ability to consider different contexts and estimate the uncertainty in the prediction. Seven models from four families of machine learning algorithms are investigated for their novel utilisation. A novel concept of utilising a Baysian Neural Network to the binary classification problem of propensity-to-pay energy bills is proposed and explored for deployment.

\end{abstract}

\section{Introduction}

{I}{n} the Australian Energy sector quarterly billing is common practice. This may be due to the economics of manually read meters that require human meter readers to take physical readings from the meter.  While smart meter technology is allowing meters to be read more frequently, there are still large numbers of manually read meters in use. Large and infrequent energy bills can present financial pressure to some customers. Advance knowledge of an individual customer's propensity to pay a bill can provide organisations with many opportunities to improve the customer experience, reduce hardship and reduce the risk of impaired cash flow or occurrence of bad debt, leading to improved outcomes for customers and the organisation. 

Predicting a customer's propensity-to-pay at an early point in the revenue cycle can help energy retailers know when to offer payment options, such as payment plans, to smooth payments and reduce customer hardship. In order to target the offers of assistance to the most vulnerable customers, it is necessary to identify at-risk customers in a timely proactive manner. In addition to creating a good customer experience, the timely receipt of payments is needed to assist organisations to remain financially viable \cite{zeng2008using, paul2012impact, baesens2003benchmarking}. With delayed bill payments, there is a risk of reduced cash flow and a rise in bad debt write-offs \cite{baesens2003benchmarking} which can in turn contribute to increased service costs for all customers \cite{paul2012impact}.

In many industries, organisations use credit scoring to understand customers' payment patterns and ability to pay \cite{anderson2007credit}. The traditional methods of credit scoring are based on statistical methods, such as logistic regression \cite{wiginton1980note, reichert1983examination, leonard1993empirical}. However, these methods rely on access to large and detailed datasets to accurately predict a customers propensity-to-pay.

In this paper, we show that predicting a customers propensity-to-pay a bill can be achieved with machine learning models while using limited amounts of de-identified data combined with publicly available Australia Bureau of Statistics\footnote{https://www.abs.gov.au} (ABS) census data. There is a growing awareness and new legislation, such as the European GDPR \footnote{https://eur-lex.europa.eu/legal-content/EN/TXT/?uri=CELEX:32016R0679} covering the amount of data collected and the ethics of using customer information. We demonstrate that by feature engineering from a limited amount of data we are able to provide useful insights without the need to unduly intrude on a customers privacy.

Machine learning models built for predicting propensity-to-pay can provide this information with higher accuracy and more certainty \cite{zeng2019user,crook2007recent,huang2004credit,tsai2010credit}. Machine learning techniques infer common rules and patterns from a training dataset \cite{selz2020electronic}. When applied to a new and unseen situation, the trained model makes predictions using the learned generalised patterns \cite{crook2007recent, bishop2006pattern}. Machine learning approaches can utilise data from multiple and disparate sources to create a single source of truth. For example, these models can leverage customers' transaction history, behavioural interactions and augmented third party data for accurate prediction. Using more data available for modelling assists these models favourably. 

Some of the commonly applied predictive machine learning models used  for structured data (i.e. tabular data) are decision tree based methods such as XGBoost \cite{chen2016xgboost}, Random Forest \cite{liaw2002classification}, Logistic Regression \cite{hosmer2013applied} and probability based methods such as Naive Bayes. Recently Deep Neural Network (DNN) \cite{glorot2010understanding} models have gained popularity because of their effectiveness in handling a large number of variables as well as categorical variables (or features) through entity embedding \cite{guo2016entity}. 

During training of machine learning models, they learn common patterns. In other words, an optimal point estimate for the parameters in model is learned from the data samples \cite{blundell2015weight, neal2012bayesian}. These point estimate parameters are then used to make prediction on the previously unknown samples. They provide a crisp decision in the form of prediction classes, for example, propensity-to-pay "yes" or "no". They do not provide uncertainty or probability of yes or no to a sample. Additionally, these models perform well when there are adequate training samples available. 

Many machine learning models fail to communicate uncertainty in regions with scarce or no data, resulting in overconfident prediction \cite{neal2012bayesian, blundell2015weight}. Without knowing the uncertainty, such overconfident predictions can lead to wrong decision making, which can cost a lot to organisations. 
Naive Bayes-based models \cite{rish2001empirical} are simple to implement and can provide uncertainty. However, Naive Bayes assumes that all features are conditionally independent, and therefore cannot capture correlations between input features \cite{rish2001empirical}. As a result, the accuracy of these models does not increase with the addition of training examples \cite{ng2002discriminative}. 

This paper presents a case-study conducted on a dataset obtained from an energy organisation and investigates machine learning models to predict if a customer is going into, or about to go into, financial hardship that is likely to affect their ability to pay their energy bills by the due date. The goal is to allow the organisation to take proactive action to assist customers and reduce the negative effects that missing payments can bring to the customer. We investigate seven models from four families of machine learning algorithms for their novel utilisation. 

The point-estimates based models are Deep Neural Network (DNN) \cite{glorot2010understanding}, Decision Tree based models (XGBoost \cite{chen2016xgboost}, Random Forest \cite{liaw2002classification} and a single Decision Tree \cite{safavian1991survey}), Logistic Regression \cite{hosmer2013applied} and Multinomial Naive Bayes (MNB) \cite{lewis1998naive}. We propose a Bayesian Neural Network (BNN) using Variational Bayes Inference \cite{kingma2013auto} to predict propensity-to-pay. By introducing the concept of Bayesian to NN, BNN adds an uncertainty estimation and model regularisation mechanism for predictions. BNN introduces probability distribution over the parameters of a traditional NN \cite{blundell2015weight}. BNN is based on the Bayes by Back-prop algorithm \cite{blundell2015weight} which estimates a variational approximation for the true posterior. This allows BNN to estimate the uncertainty by sampling from the posterior. 

The data used was a limited amount of deidentified energy bill payment history data to conduct experiments and investigate the use of machine learning in propensity-to-pay energy bills prediction. Several measures were used to evaluate the performance of models in predicting propensity-to-pay. Results show that machine learning models were able to learn from the data and predict a customer is going into, or about to go into, financial hardship. The performance of these models vary moderately under different measures. The trees in the proposed XGBoost model generated forest can be used to identify the features that contribute most in propensity-to-pay. The proposed BNN model not only achieves equivalent performance to traditional models but also incorporates a mechanism to communicate uncertainties in the underlying data distribution. BNN allows the prediction made by the model to be more reliable. 

The paper makes the following novel contributions.

\renewcommand{\labelenumii}{\labelenumi\alph{enumii}) }
 \begin{enumerate}
   \item presents an evidence-based data-driven solution to the problem of propensity-to-pay and train a number of  machine learning models for predicting propensity-to-pay energy bills.
   \item Presents how Bayesian Neural Network can be effectively applied to this problem for estimating prediction uncertainty.
    \item Shows that a machine learning model based on decision tree can be used to identify the features that contribute most in propensity-to-pay. 
  \end{enumerate}

The rest of the paper is organised as follows. Section \ref{sec:related_work} discusses related work. Problem formulation is given in Section \ref{sec:problem}. The problem solving framework is given in Section \ref{sec:framework}. Machine learning models are discussed in Section \ref{sec:models}. Empirical evaluation and result discussion are given in Section \ref{sec:evaluation}. Section \ref{sec:conclusion} concludes the paper. 

\section{Related Work}
\label{sec:related_work}
Organisations face economic and social pressure to deliver high quality services while maintaining affordable prices. Some of the efficiencies needed to remain profitable in a competitive landscape can be obtained through improvements in business efficiency and reduction in costs. Maintaining a good cash flow and reducing bad debt write-offs are ways that organisations can reduce service costs to the communities they serve and improve customer experience. 
Predicting propensity-to-pay can greatly help in this regard and utilising an automatic way for such prediction is considered co-creating value for both customers and organisations \cite{hein2019digital}.
Propensity-to-pay scoring is a common idea in many industries and has been used by industries such as banking and insurance to determine where to focus their resources.
\subsection{Traditional Methods}

Traditional methods include use of human judgement based on previous experience \cite{henley1996ak}, discriminant analysis and linear regression \cite{durand1941risk, srinivasan1987credit}. However, discriminant analysis and linear regression are subjected to conceptualisation problems \cite{eisenbeis1978problems, berry1994estimating}, therefore logistic regression is more commonly used \cite{wiginton1980note, reichert1983examination, leonard1993empirical}. 
Survey based methods have also been used. A cross-sectional survey study was conducted in 11 major towns in Uganda to understand the perceptions of water utility customers that influence their bill payment behaviour and found that service value and customer satisfaction influence bill payment behaviour \cite{kayaga2004bill}. Recently, a discrete choice model \cite{berry1994estimating} was used to estimate willingness-to-pay for energy efficiency \cite{collins2018willingness}. 

\subsection{Machine learning Methods}
Machine learning models and techniques can provide solutions with higher accuracy with more certainty than traditional models \cite{west2000neural, tsai2008using, atiya2001bankruptcy, khashman2010neural,bellotti2009support, kim2012corporate, chen2014credit}. They have capacity to handle large and multiple data sources as well as make predictions without human bias, and can be deployable as part of fully automated implementation. Machine learning has been used successfully in diverse applications across industries. There exists a handful of studies that have used machine learning to understand the customers' bill or credit payment behaviour. Some of them use Decision Trees (DT) \cite{hongxia2010enterprise}, K-Nearest Neighbour (KNN) \cite{fajrin2018credit, henley1996ak, li2009hybrid}, Support Vector Machine (SVM) \cite{bellotti2009support, kim2012corporate, chen2014credit} and Neural Networks (NN) \cite{west2000neural, tsai2008using, atiya2001bankruptcy, khashman2010neural}. 

DT presents the knowledge learned by the model comprehensibly such that people can understand the decision making process in an intuitive manner without the requirement of special training. The C4.5 decision tree has been reported to provide better predictive accuracy than most of the statistical methods \cite{quinlan2014c4}. SVM can classify non-linearly separable data sets by using different kernel tricks, and therefore can improve accuracy. KNN models are computationally efficient. NN models can deal with complicated prediction problems containing a large number of variables. 

A survey of applying machine learning techniques to credit rating reported that the predictive accuracy of classifiers varies in datasets \cite{wang2015survey}. They observed, for example, KNN classifier provided the best accuracy on one dataset and worst accuracy on another dataset. However, other similar studies  \cite{baesens2003benchmarking} did not find significant difference in the performance of classification models. They observed that the least-squares support vector machine models (LS-SVMs) with radial basis (RBF)  kernel and NN classifiers performed marginally better than simple linear models due to the data being weakly non-linear. In summary, the performance of a machine learning model depends on the nature and characteristics of the dataset. 

Recently, Allina Health partnered with Health Catalyst reported a case study of propensity-to-pay prediction of medical bills\footnote{https://downloads.healthcatalyst.com/wp-content/uploads/2019/04/Allina\_Propensity-to-Pay\_case-study.pdf}. The initiative has been reported to result in \$2 million increase in overall bill collections in just one year. A Random Forest model was implemented where the decision trees in the forest were built using Gini impurity index.

\subsubsection{Deep Learning}
Recently Deep Neural Network (DNN) \cite{glorot2010understanding} models have emerged as winners amongst machine learning models in the areas of computer vision and natural language processing. They have been reported to perform excellently for data with a large number of variables. These models have also been successfully applied in time-series data. Deep learning models have been highly used in Kaggle competitions\footnote{Kaggle offers free tools to run academic and recruiting machine learning competitions. link: https://www.kaggle.com/competitions}, e.g., 1st place was achieved to predict the taxi destination using a trajectory of GPS points and timestamps \cite{de2015artificial}, 3rd place was achieved to predict future sales using time series data from a chain of stores \cite{guo2016entity}. However, DNN models have received less attention dealing with tabular (i.e. relational) data even though tabular data is very common and highly valued in many Data Science and analytics projects. This is partly because DNN has historically tended to overfit the training data. Recent works show that regularisation methods, such as random dropout \cite{srivastava2014dropout} and L2 normalisation, can reduce overfitting, and therefore DNN is gaining attention in tabular data. It will be interesting to explore their potential in learning patterns from tabular data.

\subsubsection{Uncertainty in prediction}
Machine learning models (e.g., NN and many other symbolic models such as DT, Support Vector Machine (SVM), Logistic Regression, etc.) use single-point estimates as parameter values (i.e., weights and biases) during the class prediction \cite{pawlowski2017implicit}. It is not justifiable to use single-point estimates if we consider a  probability  theory  perspective \cite{shridhar2019comprehensive}. The prediction applications will be benefited by adding uncertainty in the prediction, and an outcome to be undefined if the model is uncertain, instead of forcing a model to choose one of the classes. Naive Bayes-based models \cite{rish2001empirical} can provide uncertainty. However, it assumes that all features are conditionally independent, and therefore it cannot capture correlations between input features \cite{rish2001empirical}. As a result, the accuracy of these models does not increase with the addition of training examples \cite{ng2002discriminative}. 

A Bayesian neural network (BNN) allows a neural network to learn and present this uncertainty with a prior distribution on its weights \cite{neal2012bayesian}. Because the average value of a parameter is computed across many models during training, these models achieve a regularisation effect in the network that prevent it from overfitting \cite{shridhar2019comprehensive}. Consequently, BNN models can learn from small datasets. A number of image classification applications have successfully used BNN based models \cite{shridhar2019comprehensive, blundell2015weight, gal2016dropout} 
 However, a BNN model has not been investigated for tabular data, partly because DNNs are not well utilised for tabular data until recently. With these potential characteristics of BNN, we propose to investigate and apply a BNN model on the tabular data obtained for the propensity-to-pay problem.

To our best of knowledge, machine learning models have not been utilised in propensity-to-pay prediction of energy bills. We propose to investigate and apply various machine learning models (including DNN and BNN) to the propensity-to-pay problem in which data collection is constrained by the consumer privacy objectives and law while not all of the data collected should be used in machine learning. We propose to perform feature engineering from a limited amount of data without the need to unduly intrude on a customers privacy.

\section{Problem Formulation}
\label{sec:problem}
A utility company with a focus on innovation \cite{wiesbock2019digital} and customer affordability is usually interested in exploring data-driven decision making to assist in supporting customers through financial hardship, in particular, through early detection and assistance with the aim of reducing negative impacts on the customers. We propose to investigate the capability and benefits of machine learning models (from four families, i.e., Decision Tree, Deep Learning, Bayesian and Logistic Regression) in predicting propensity-to-pay. We explore the uncertainty around the creation of machine learning models that are able to predict residential customers entering financial hardship, resulting in a reduction in their ability to pay energy bills. We assess whether a model produces results that could be used within the business to trigger a proactive harm-reducing actions. 

The propensity-to-pay problem can be framed as a question: \emph{how likely will a customer pay their bill in time}. For the purpose of this work to create a clear definition; the ideal time to make this prediction is taken as the point at which a bill was generated. A solution is focused around the energy bills being generated and the likelihood that the bill will be paid on-time by the customer. Predicting propensity-to-pay for a customer can be difficult due to several reasons such as access difficulty to predictable features due to privacy issues, lack of predictable features in the record, use of a suitable prediction algorithm, etc. Each customer was represented by an array of features using the context of a bill such as income, age, address of a customer, amount of the bill, payment mode of the bill, remoteness of the customer's living area, etc. Determining the likelihood of various features' contribution to prediction is difficult especially when all these features are observed in different times, locations and context in the organisation. Overconfident and wrong predictions can result in inefficient resource allocation and utilisation, e.g. spending staff time offering assistance to customers who don't need it while missing customers who need assistance. A machine learning method should consider different contexts and be able to model the uncertainty in the propensity-to-pay prediction problem. 

\section{The Proposed Solution}
\label{sec:framework}

\begin{figure}[htb!]
	\centering
	\includegraphics[scale=0.5]{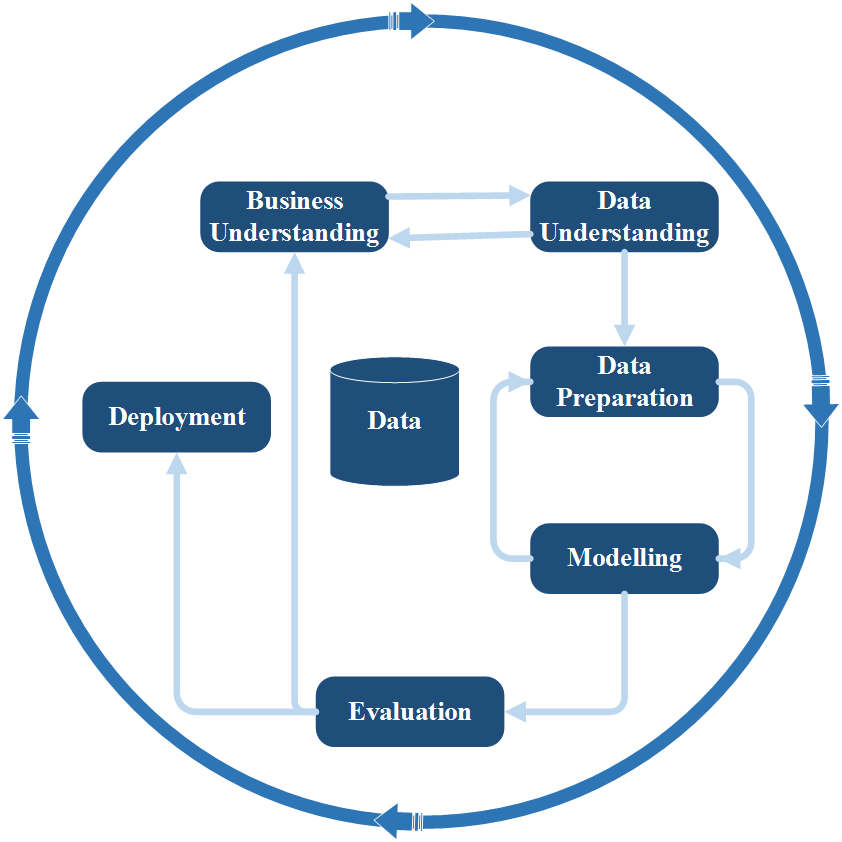}
	\caption{The proposed machine learning framework based on CRISP-DM (modified figure from \protect \cite{wirth2000crisp})}
	\label{fig:crisp_dm}
\end{figure}

We present the proposed machine learning framework that follows \emph{Cross-industry Standard Process for Data Mining} (CRISP-DM) methodology \cite{wirth2000crisp}, as shown in Figure \ref{fig:crisp_dm}. The detail of the framework is as follows.

\subsection{Business Understanding}
The  business  objectives  and  requirements  are discussed in Section \ref{sec:problem}. The anticipated output is to assess whether a model produces results that could be used within the business to trigger a preventative action. The proposed solution should be able to:
(a) estimate the uncertainty of the prediction;
(b) identify influencing features;
(c) identify the predictive power of currently available features; and 
(d) identify the suitable models.

\subsection{Data Understanding}

\subsubsection{Data Acquisition} 
Data was identified by business experts who are familiar with organisational data-sets and the business. The datasets that capture information in relation to accounts, bills, debt history, premises, payments, and segmentation were identified as useful source. The relevant data was extracted and deidentified to be ready for modelling. In addition publicly available data was identified as being potentially useful.  To ensure customer privacy, identifying information was removed and a unique identifier was added to each observation. The data used was also reviewed to ensure that the minimum amount of customer information was used and the used information was deidentified to a high standard while still maintaining functionality this retained the ability to be useful. Some of the examples are replacing the customer name with a unique index, replacing address with mesh block (A mesh block is a geographic grouping of several properties; and it is the smallest geographical unit for which Census data are available. Mesh blocks cover the whole of Australia without gaps or overlaps.\footnote{https://www.abs.gov.au/websitedbs/censushome.nsf/home/meshblockcounts}) and replacing exact age with a range. This was done to ensure that the  privacy of customers is protected to the highest possible standard. 

\subsubsection{Data Exploration}
Through this stage all variables were analysed and assessed for relevance. Each dataset was analysed independently to understand the format, type, volume, uniqueness and distribution. The variable meanings were clearly understood. This step identified variables that had many missing values, or corrupt values or outlier values. For example, we used variance and frequency distribution to identify and remove outlier values, we checked against expected range and values, as guided by business experts, to identify corrupt values. 

\subsubsection{Assumptions and Limitations} 
Some data complexity was removed in the dataset by considering the focus of the business problem and data availability. For example, this problem was focused on residential accounts, hence commercial customers data was not included.  The complex account with bespoke billing arrangements and large bills were excluded from the dataset as large bills are assumed to relate more closely to a small business customers than an standard residential accounts.

\subsection{Data Preparation}
The data sets from internal and external sources were wrangled together to create a single main source for feeding the models. Considerable time was spent to pre-processing the data for analysis. Pre-processing is an important step and the phrase ``garbage in, garbage out'' is true for any machine learning project \cite{pyle1999data}. We applied data cleaning, integration, transformation, reduction and discretisation in the pre-processing stage.  Some examples are as follows. String values were converted to lower case. A date was converted to several variables breaking the information on day, week, week day, month and year. Outliers and instances with missing values were removed. Irrelevant variables were removed after consulting with business experts. Redundant variables were removed based on the analysis of correlation coefficient. 
Continuous variables were normalised for some models and discretised for others. Categorical variables were encoded with numeric values. 
Variables were carefully considered and chosen for each model. A detailed  description of data is given in Section \ref{sec:data}.

\subsection{Modelling}

Seven models from four family of machine learning algorithms are chosen for investigation, namely decision tree family (XGBoost, Random Forest, and simple Decision Tree), neural network (Bayesian Neural Network and Deep Neural Network), linear model (Logistic Regression) and Bayesian (Multinomial Naive Bayes). The decision tree family is chosen to make the model explainable and to intuitively identify influential features. Bayesian Neural Network (BNN) was chosen to estimate the uncertainty around the prediction and Deep Neural Network (DNN) is used to  benchmark the BNN to see the effect of presenting uncertainty on the datset. Multinomial Naive Bayes (MNB) is used to show that accuracy of BNN and DNN is much higher than Naive Bayes. Finally, as logistic regression is commonly used in business domain, it is used to benchmark all the models. It is also used to test the common conjecture stating that ``the simple model works well". All the models are to be assessed against the business objective that their accuracy and recall are high, they can provide uncertainty of prediction and their predictions are interpretable. Detailed description of machine learning models are given in Section \ref{sec:models}.


\subsection{Evaluation}
Performance of each model is reviewed considering their strengths and weaknesses (e.g. comprehension vs accuracy). Each model was checked against the business objectives, considering what and how can the model be implemented and deployed. Empirical analysis has been conducted to identify the data suitability, features importance, best model selection and the possibility to deploy this model in practice in everyday decision-making. The ability of a model to be rapidly deployed and modified in the future, all at an economical cost, is also a key business requirement. 
Detailed empirical evaluation and analysis are given in Section \ref{sec:evaluation}.

\subsection{Deployment}
Internal reports covering business and technical matters have been prepared to inform key stakeholders within the organisation. These reports will assist with detailed cost/benefit analysis around the deployment of these machine learning models in an operational context. It is anticipated that further model development may be required along with a comprehensive review of these processes and testing of assumptions. The business would require a  detailed benefits-analysis and better understanding of costs to implement this technology based on the findings from building of these machine learning algorithms in this phase to predict propensity-to-pay. Key deployment considerations would be reviewing assumptions, data integrity and availability, system capabilities, model governance, ongoing cost/resourcing and skill-set required to monitor and develop model accuracy and continuity.

\section{Machine Learning Models}
\label{sec:models}

Let $\mathbf{x} = (x_1,\dots x_n)$ be a vector representing an instance (i.e., context of a bill in the propensity-to-pay problem such as income of the customer, amount of bill, remoteness of the customer's living area, etc.) with $n$ features and a class label $C = \{0, 1\}$ where $C$ is a set of binary classes with class 1 representing bill paid in time and class 0 representing bill that was/will not be paid on time. The classification task is to assign an instance to a class $C_k \in C$ based on the feature vector $\mathbf{x}$, i.e. finding $p(C_k|\mathbf{x})$. We want to learn a model with parameters $\theta$ using training data (i.e., the historical payment data) $D$ that reasonably approximates $p(C_k|\mathbf{x}) \approx p(C_k|\mathbf{x}, \theta)$. Next, we discuss a number of models used in the process of predicting a customer's propensity-to-pay for a bill.  

\subsection{Neural Network Based Models}

\subsubsection{Deep Neural Network} 
In the last decade, Deep Neural Network (DNN) \cite{glorot2010understanding} models have become popular in computer vision and natural language processing. Some of the reasons are their capability to deal with a large number of variables, their ability to achieve better accuracy and their non-reliance on domain-specific feature engineering during training. However, it has received far less attention to solve problems that can be represented as in tabular (i.e. relational) data even though tabular data is very common and highly valued in many Data Science and analytics projects. 

\begin{figure}
	\centering
	\includegraphics[scale=0.35]{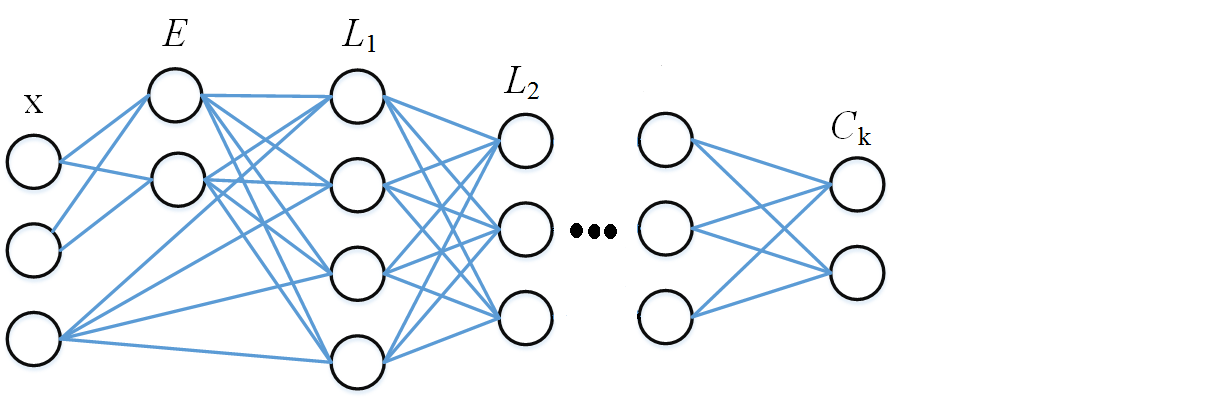}
	\caption{A Deep Neural Network Model}
	\label{fig:dnn}
\end{figure}

We propose to apply a DNN model\footnote{Code available at https://github.com/mdabashar/Propensity-to-Pay} with embedding training to tabular data to learn propensity-to-pay prediction. The architecture of the proposed model is given in Figure \ref{fig:dnn}. In this architecture, $x \in X$ is input samples, $E$ is embedding layer, $L_1$ and $L_2$ are hidden layers and $C_k$ is prediction in the final (or classification) layer. Each layer consists of a number of nodes or neurons. When $x \in X$ is a categorical variable it is connected to an embedding layer. Embedding layer is equivalent to an extra layer on top of the encoded input so that each category can be represented as a vector of floating point numbers\cite{guo2016entity}. It allows the network to learn the best representation for each category during training by capturing rich relationships between different values of a category. For example, there may be patterns for post codes that are geographically near each other, or for post codes that are similar in socio-economic status. An embedding layer quantifies semantic similarities between values of a category based on their distributional property by mapping co-occurring values close to each other in an Euclidean space.  

In (unstructured) text data, skip-gram and continuous Bag-of-Words are two popular models for word embedding \cite{mikolov2013distributed}. Let the current value in a sample be $e_i$ and the other values be context entities $C$. The continuous bag-of-words model predicts the current value $e_i$ from the context entities $C$, i.e. $p(e_i|C)$. The skip-gram model uses the current entity $e_i$ to predict the context entities $C$, i.e. $p(C|e_i)$. The objective of embedding training is to find an entity embedding that maximises $p(e_i|C)$ or $p(C|e_i)$ over a dataset. In each step of training, each entity is either (a) pulled closer to the entities that co-occur with it or (b) pushed away from all the entities that do not co-occur with it. 

We propose to apply entity embedding in DNN by adding an extra layer on top of encoded inputs \cite{guo2016entity,kim2014convolutional,bashar2018cnn,bashar2020regularising}. At the end of the training, embedding brings closer together values in a category, in a training dataset, that are not only explicitly co-occurring but also the values that implicitly co-occur. For example, if $e_1$ explicitly co-occurs with $e_2$ and $e_2$ explicitly co-occurs with $e_3$, then embedding can bring closer not only $e_1$ to $e_2$, but also $e_1$ to $e_3$. Because an embedding captures richer relationships and complexities than the raw values in a category, the learned embeddings for categorical variables (e.g. product, store id, or post code that are commonly used in a business) can also be used for other models. 

Output of the embedding layers and continuous variables ($x \in X$) are normalised and connected to fully connected hidden layer $L_1$ (Figure \ref{fig:dnn}). Output of $L_1$ is fed to fully connected hidden layer $L_2$. Finally, output of $L_2$ is passed to fully connected final layer that predicts the class $C_k$ of the given input instance.

Each node in DNN has two parts: a linear part and a nonlinear part. 
The linear part of the node is a function $f$ that maps the input samples $\mathbf{X}$ into an intermediate representation $Z$, i.e. $Z = f(\mathbf{X}, W, b) = \mathbf{X}.W + b$, where $W$ is a weight matrix and $b$ is a bias vector. A non linear function $g$ is then applied to $Z$ to get the outcome $L$ of the node, i.e. $L = g(Z)$. A set of nodes together constitute a layer. In this paper, we empirically use two fully connected hidden layers (Figure \ref{fig:dnn}). 
\begin{align*}
Z_1 &= f(\mathbf{X}, W_1, b_1)\\
L_1 &= g(Z_1)\\
Z_2 &= f(L_1, W_2, b_2)\\
L_2 &= g(Z_2)
\end{align*}
In the hidden layers we use ReLu as the nonlinear function $g$, i.e. $g(z \in Z) = ReLu(z) = max(z,0)$. In the final layer, we want to get two probability distributions for two classes. Therefore, we use a softmax function as the activation of the output layer. 
\begin{align*}
Z_o &= f(L_2, W_3, b_3)\\
p(C_k|\mathbf{x})\approx g(Z_o) &= softmax(Z_o) = \frac{e^{Z_0}}{\sum e^{Z_0}}
\end{align*}

\subsubsection{Bayesian Neural Network}
In many real-world applications including propensity-to-pay, there is a benefit in understanding the confidence or uncertainty the model has about the predictions that it makes. It may be useful for a machine learning-based decision-making system to not act  when the prediction has high uncertainty\footnote{https://bit.ly/2AsHclo}.

Deep neural networks (and many other machine learning models) are trained to assign a class from the predetermined classes to a test sample even if the sample is completely unrelated to the data used for network training. For example, consider a network that has been trained as a binary  ``cat vs dog" classifier. In the situation where it receives a test image of a person to be classified, it would either classify it as a cat or as a dog. It is unable to communicate that the person does not closely resemble a dog or a cat. Since the output of the final layer (softmax) is interpreted as the probabilities,  the network will always produce an output with the highest probability even if that probability is very low. There will always be a class with the maximum value for a test sample. The binary DNN classifier does not have a way to communicate the model's uncertainty about the data it has not been trained to handle. The goal of BNN is to enable the network to communicate this uncertainty information. 

We propose to apply a BNN model\footnote{Code available at https://github.com/mdabashar/Propensity-to-Pay} to estimate uncertainty when predicting propensity-to-pay. When making business decisions the cost of the action or inaction can be accurately considered for each prediction based on the corresponding uncertainty of the prediction. Implementation of BNN in propensity-to-pay prediction can result in more efficient resource allocation for the business. 

\subsubsection{Uncertainty prediction}
\begin{figure}
	\centering
	\includegraphics[scale=0.6]{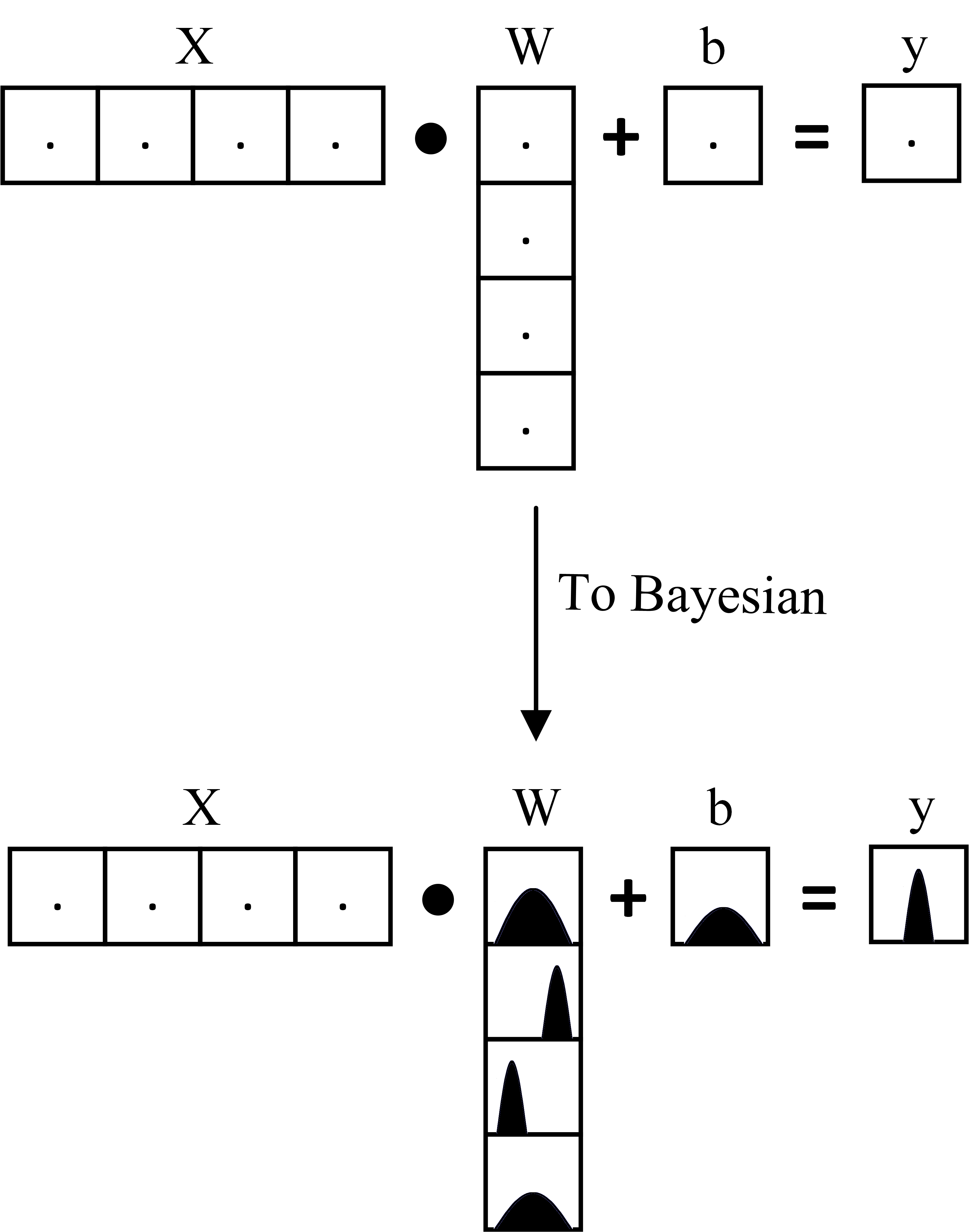}
	\caption{Converting Linear Regression to Bayesian Linear Regression}
	\label{fig:to_bayesian}
\end{figure}

In traditional NNs, each parameter (i.e., weights and biases) has a fixed value that determines how an input is transformed into an output.
In BNN, a probability is attached to each parameter \cite{blundell2015weight}. For simplicity, we can say that each parameter is turned into a random variable. Figure \ref{fig:to_bayesian} illustrates how to convert a single layer NN without nonlinearity (i.e., equivalent to Linear Regression) into a single layer BNN without nonlinearity (i.e., equivalent to Bayesian Linear Regression).

Formally, all parameters $\theta$ are transformed into random variables $\Theta$ and some \emph{a-priory} probability distributions $p(\Theta)$ are assigned. The training data $D$ is then used to update the probability distributions $p(\Theta|D)$ through Bayes theorem \cite{downey2012bayes} as follows. 
\[p(\Theta|D) = \frac{p(D|\Theta)p(\Theta)}{p(D)}\]
where $p(D|\Theta)$ is the likelihood of $\Theta$ to describe data $D$.

However, estimating the posterior $p(\Theta|D)$ is difficult because the marginal probability distribution $p(D) = \sum_\Theta p(D|\Theta)$ can not be estimated as each random variable in $\Theta$ can have values from negative infinity to positive infinity and there can be millions of random variables in $\Theta$. Therefore, sampling based approaches such as Markov-Chain-Monte-Carlo (MCMC) \cite{andrieu2003introduction} can be used to address this problem. However, MCMC is very slow and will take an unreasonably long time for the large number of random variables such as in BNN \cite{bardenet2017markov}. Consequently, the gradient
descent-based Variational Bayes \cite{kingma2013auto} has been commonly used to approximate the posteriors for BNN. The Variational Bayes method uses the Evidence Lower Bound (ELBO) when approximating the posterior so that the distance between the true posterior and approximation can be found during optimisation.

A random variable gives a different value each time when accessed. Each time, the obtained value depends on the random variable's associated probability distribution. The higher variance the associated probability distribution has, the more uncertainty there is in regards to its produced value, as the random variable could provide any value as per the variance of the probability distribution. 

The process of getting a value from a random variable is called sampling. To classify an instance, the BNN is run multiple times (forward pass), each time the network samples a new set $\theta$ of parameter values (weights and biases). Instead of a single value $p(C_k|\mathbf{x}, \theta)$ for a class, multiple values are obtained, one value for each run. The set of values represents a probability distribution over the classes. Therefore, confidence and uncertainty for the class $C_k$ of test instance $\mathbf{x}$ can be determined. If the test instance comes from a data distribution that the network has not learned enough, the uncertainty will be high. It can be interpreted as the network expressing uncertainty about the prediction. 



\subsection{Tree Based Models}
Tree based models are one of the most popular in generating predictions from structured data. Especially, Gradient Boosted Trees (e.g. XGBoost) and Random Forests have been successful in many Kaggle competitions \cite{chen2016xgboost}. We briefly discuss the basic decision tree model and two popular tree ensemble models, random forests and gradient tree boosting XGBoost. 

\subsubsection{Decision Tree}
Decision Tree (DT) \cite{safavian1991survey} builds a classification model in the form of a tree structure comprised of decision nodes and leaf nodes. An internal node is called a decision node and it has branches. Each leaf node represents a class concluded from a series of decision nodes following a decision path. It progressively breaks down a dataset into smaller subsets while adding a node to incrementally build the associated decision tree. 

Information gain and Gini index are the common measures to evaluate a decision node (i.e. to determine which feature to use as a node) when incrementally building the tree. Gini index is a measure of impurity or homogeneity in the data. Gini index for a given feature $x_i$ is estimated as follows.
\[Gini (x_i) = \sum_{j=1}^m p_j \left(1 - \sum_{k=1}^K p_k^2\right)\]
where $m$ is the number of branches for split for feature $x_i$, $p_j$ is the probability of branch $j$, $K$ is the number of classes and $p_k$ is the probability of $k$th class in each branch $j$. Gini index gives value between 0 and 1, 0 being perfectly homogeneous and 1 being maximum impurity or heterogeneous. For example, for a given feature $x_i$, if most of the samples are homogeneous (i.e. belong to the same class), its Gini index will be low, else it will be high. 

\subsubsection{Random Forest}
A single decision tree usually shows a high variance. To address this problem, Random Forest (RF) \cite{liaw2002classification} establishes a committee of $N$ decision trees. It is an ensemble learning method for classification that constructs multiple decision trees during training and outputs the class that is the mode of the classes predicted by individual trees. 
RF builds a decision tree by randomly selecting $n$ samples from the training data (also called bootstrap sampling), where $n$ is smaller than the number of total samples in the training data. It repeats this process $N$ times to build $N$ trees in the forest. 

\subsubsection{XGBoost}
XGBoost (XGB) (gradient boosting decision tree) \cite{chen2016xgboost} is one of the most popular machine learning algorithms for tabular data, especially with a small number of variables. XGB is an ensemble tree (committee) based model. Like any other boosting methods, XGB builds the model in a stage-wise fashion. At each stage $m$, $1 \leq m \leq M$, it assumes that there is an imperfect model $F_m$. Therefore, it construct a new model $F_{m+1}$ that adds an error estimator $h$ to $F_m$ to improve it, i.e. $F_{m+1}(\mathbf{x}) = F_m(\mathbf{x}) + h(\mathbf{x})$. In an ideal situation $F_{m+1}(\mathbf{x}) = F_m(\mathbf{x}) + h(\mathbf{x}) = C_k$, or $h(\mathbf{x}) = C_k - F_m(\mathbf{x})$. Gradient boosting fits $h(\mathbf{x})$ to the error $C_k - F_m(\mathbf{x})$. That is, each model $F_{m+1}$ aims to correct the errors of the previous model $F_m$. The error $C_k - F_m(\mathbf{x})$ is a negative gradient (with respect to $F(\mathbf{x})$) of the loss function $\frac{1}{2} (C_k - F_m(\mathbf{x}))^2$, which turns gradient boosting into a gradient descent optimisation.

\subsection{Multinomial Naive Bayes}
A family of simple probabilistic classifiers based on Bayes theorem with independence assumptions (naive) between the features are known as naive Bayes classifiers \cite{mccallum1998comparison}. In multinomial naive Bayes, each feature vector represents the frequencies with which certain features have been generated by a multinomial (i.e. $p_1, \dots, p_n$) where $p_i$ is the probability that $i$th feature occurs.

\subsection{Logistic Regression}
Logistic Regression (LR) \cite{hosmer2013applied} is a statistical model that  uses a logistic function to map the values from independent variables to the values of a binary dependent variable. Logistic regression is an established method that has been successfully used in various fields for predicting binary outcome such as medical science, social science and economics.

\section{Empirical Evaluation}
\label{sec:evaluation}
This section details the performance of machine learning models in predicting propensity-to-pay energy bills. The models are implemented using python on Jupyter\footnote{https://jupyter.org/} notebooks. Code versions are managed using a git\footnote{https://git-scm.com/} repository. 

\subsection{Data Collection: Variables selection}
\label{sec:data}
To implement and test the machine learning models, we used the data collected from a Queensland utility company. Data was shared using secure transfer mechanisms to comply with legal, ethical and business privacy requirements. To predict if a customer is likely to pay energy bills on time, we need to analyse the features (or variables) that influence the payments made in the past. We combine the historical (and labelled) data from a number of sources to train machine learning models to determine how variables have contributed to payment behaviour in the past, thus to predict how a customer with these features will respond in the future. 

After carefully considering variables from internal and external sources, we selected 34 independent variables to using in predicting the dependent variable of whether a customer will pay a given bill in time. Out of the 34 independent variables, eight came from the external data source of Australian Bureau of Statistics (ABS). Independent variables cover customer age, bill issue and due dates, bill sending method, physical remoteness from a local population centre, income and wealth grouping, bill duration and account age at an individual level. Besides, these variables cover median data in the SA1 group of ABS data for household income, household size, person per bedroom, weekly income, weekly rent and weekly mortgage.  




\subsection{Data preprocessing: Feature engineering}
Several variables contain categorical values of string nature. We converted all strings to lower cases, dropped null values, dropped variables with highly imbalanced categories (e.g. 99\% and 1\%), and removed outliers based on variance and frequency distribution.
We split the bill due date into multiple categorical variables such as year, month, week, day of week, and day of month. This splitting is in line with other research where deep learning was successfully used with time series data \footnote{https://www.fast.ai/2018/04/29/categorical-embeddings/}.  

\subsection{Data distribution for modelling and testing}
After all the preprocessing and feature engineering, the sample data-set selected for modelling included a total of 5.05 millions of instances of bills issued. Out of that, 2.1 million (41.63\%) are negative instances (did not pay bill on time) and 2.95 million (58.37\%) are positive instances (paid bill on time). We note that this is not representative of real world ratios, sample data was selected to produce positive/negative ratios that are close to being a balanced data set.

To apply machine learning models to time-series data, it is necessary to choose validation and test sets that are a continuous selection with the latest available dates in the data. Choosing a random subset of the data is not representative of most business use cases because we are using historical data to build a model for use in the future\footnote{https://www.fast.ai/2017/11/13/validation-sets/}. We split the data into 80\% for training set and 20\% for testing set using a continuous selection with the latest available dates. The training set contains a total of 4.04 millions of instances and the testing set contains 1.01 millions of instances. Training set has 1.683 million (41.66\%) negative instances and 2.357 million (58.343\%) positive instances. Testing set has 0.419 million (41.521\%) negative instances and 0.591 million (58.479\%) positive instances. We observed that for this dataset oversampling improves accuracy of the models. Therefore, we used oversampling to stratify the training set but did not stratify the validation and test sets to make sure that the performance reflects the real world scenario. 

\subsection{Machine Learning Models: Configuration and Training}
The following models were implemented for experiments. 

BNN: We use Pyro\footnote{https://pyro.ai/} and PyTorch\footnote{https://pytorch.org/} to implement the BNN model. We use a standard Neural Networkk of one hidden layer with 1024 units. We initialise the parameters of random variables with a normal distribution. We use the loss function ELBO and mini batch for optimization. We sample $\theta$ 100 times for estimating uncertainty of prediction for each test sample $\mathbf{x}$. 

DNN: We use the deep learning framework FastAI\footnote{https://www.fast.ai/} to implement DNN model. We use two fully connected layers (200 and 100 neurons) on the top of the embedding layer for categorical features and  directly on the top of continuous features. Each fully connected layer uses ReLU activation function. The output layer contains one neuron with sigmoid activation function. No dropout is used as it did not improve the result.

Traditional ML Models: We implement Random Forest (RF),  Decision Three (DT), Multinomial Naive Bayes (MNB) and Logistic Regression (LR) using Scikit-learn\footnote{https://scikit-learn.org/}. 
For RF, we assigned max depth to 5, number of estimators (trees) to 200, splitting criterion to `gini', minimum samples required to be at a leaf to 1 and minimum samples required to split to 2. Other hyparameters are set to default. 
For DT, we assigned max depth to 5, splitting criterion to `entropy', splitter to `best', minimum samples required to be at a leaf to 1 and minimum samples required to split to 2. Other hyparameters are set to default. 
For MNB, we assigned alpha to 0.05, class prior to None and  fit prior to True.  
For LR, we assigned inverse of regularization strength (C) to 1, maximum iteration to 100, penalty to 12 and tolerance for stopping criteria to 0.0001. Other hyparameters are set to default. We used xgboost python\footnote{https://machinelearningmastery.com/xgboost-with-python/} and Scikit-learn to implement XGB model. We assigned max depth to 5, number of estimators to 300, objective function to `binary:logistic', learning rate to 0.1, subsample to 1 and minimum child weight to 1. Other hyparameters are set to default. 

Hyperparameters are tuned based on cross-validation. Hyperparameters of BNN and DNN are manually tuned, while that of RF, XGB, DT, LR and MNB are automatically tuned using GridSearch from Scikit-learn. 

\subsection{Evaluation Measures}
We used six standard evaluation measures of classification performance: Accuracy, Precision, Recall, F$_1$ Score, Cohen Kappa (CK) Score and Area Under Curve (AUC). We also report True Positive (TP), True Negative (TN), False Positive (FP) and False Negative (FN) values to report the detailed performance of models. A description of these measures is given in appendix A. For a class-wise breakdown result, we provide average (avg), micro average (micro avg) and weighted average (weighted avg). Micro avg and weighted avg are used to check that models can perform well in both minority and majority classes \cite{sokolova2009systematic}.  

\subsection{Prediction performance}
\label{sec:results}
Different models have various strengths in handling TP, TN, FP and FN and can perform differently to a majority (or minority) class. Therefore, a number of measures (e.g. accuracy, recall, precision, etc.) are used to identify performance of a model on different conditions. In other words, these measures can identify the models that are best in providing required accuracy for TP, TN, FP, or FN in a given application. Table \ref{tab:models_performance} reports the propensity-to-pay prediction performance of seven machine learning models discussed in Section \ref{sec:models}. To show how the models perform for each class (Yes and No), a class-wise breakdown of the results of these models is provided in Table \ref{tab:class_wise_models_performance}. Results show that there is no significant difference between the performance of most of the models. There is no single method that appears as best under all the measures. In general, DNN, RF and XGB achieved the best results under different measures. For instance, RF produces best performance under TP and FN (i.e. correctly predicting who will pay the bill in time); whereas DNN performs best on TN \& FP (i.e. correctly predicting who will miss paying the bill in time). However, as discussed in the paper, expectantly, BNN gives a better or closer  performance to other models (i.e. XGB, RF and DNN) and yields the high quality performance under a specific measure (e.g. Accuracy, Precision, Recall, etc.). 

XGB, RF and DNN have achieved best performance at least on one measure. A comparison shows that XGB gives best performance in Accuracy, Kappa Score and AUC. RF gives best performance in Recall and F$_1$-score. DNN gives best performance in Precision. DNN scores for highest True Negative (will not pay bill in time) and lowest False Positive. RF scores for highest True Positive (will pay bill in time) and lowest False Negative. These results indicate that DNN can provide a performance close to popular machine learning models. Particularly, DNN achieves the best performance in identifying customers who are prone to fail to pay a bill on time. This information can be used by organisations to take proactive action to correctly estimate cash-flow and implement proactive harm reduction measures to assist these customers in at an early stage.  

It is to be noted that in this investigation, we focused our attention on individual models and understanding the problem domain for knowing the feature importance and characteristics of the customers in two classes. An ensemble of DNN, RF and XBG may yield a better prediction performnace. Our future work will carry out this investigation. 

A comparison of BNN and DNN in Tables \ref{tab:models_performance} and \ref{tab:class_wise_models_performance} shows that if we take an average on Accuracy, Precision, Recall, $F_1$ score, Cohen Kappa and Area Under Curve, BNN perform better than DNN. BNN performs better than DNN in Recall by 33.22\% and F$_1$-score by 9.13\%. Performance of BNN is very close to DNN in Accuracy and AUC. These results imply that we can use BNN without compromising the accuracy performance of DNN, with the added benefit of uncertainty estimate of the underlying data distribution. 

The advantage of using BNN becomes apparent when it is compared with MNB. MNB can estimate uncertainty but its prediction performance is very poor when compared with other models. MNB assumes features are conditionally independent, but the features in datasets such as propensity-to-pay are not conditionally independent and can be non-linearly related. However, BNN does not assume conditional independence of features and it can handle non-linear relations. Therefore, BNN performs significantly better than MNB. 

Organisations can realise many benefits by targeting their assistance efforts on customers who are best placed to be advantaged from relevant assistance programs, without suffering potential negative side effects from incorrectly targeted programs and offers. We note that all the model's outputs include both false positives and negatives, the probability aspect allows the business owners to make decisions based on the probability from the model combined with expected costs of false negative or positive. BNN is able to provide organisations and business leaders with a well-informed decision-making. 

\begin{table*}[htbp]
	\centering
	\scriptsize
	\caption{Model Performance Comparison}
	\begin{tabular}{lccccccc}
		\toprule
		Measure & BNN   & DNN   & RF & XGB & DT & LR & MNB \\
		\midrule
		TP & 466683 & 350480 & \textbf{483294} & 446086 & 421342 & 431324 & 320401 \\
		TN & 196790 & \textbf{310798} & 194932 & 244870 & 249252 & 223462 & 222867 \\
		FP & 222578 & \textbf{108570} & 224436 & 174498 & 170116 & 195906 & 196501 \\
		FN & 123952 & 240155 & \textbf{107341} & 144549 & 169293 & 159311 & 270234 \\
		Accuracy & 0.657 & 0.655 & 0.672 & \textbf{0.684} & 0.664 & 0.648 & 0.538 \\
		Precision & 0.677 & \textbf{0.763} & 0.683 & 0.719 & 0.712 & 0.688 & 0.620 \\
		Recall & 0.790 & 0.593 & \textbf{0.818} & 0.755 & 0.713 & 0.730 & 0.542 \\
		F$_1$-score & 0.729 & 0.668 & \textbf{0.744} & 0.737 & 0.713 & 0.708 & 0.579 \\
		Kappa Score & 0.269 & 0.320 & 0.295 & \textbf{0.343} & 0.308 & 0.266 & 0.072 \\
		AUC  & 0.630 & 0.667 & 0.642 & \textbf{0.670} & 0.654 & 0.632 & 0.537 \\
		\bottomrule
	\end{tabular}%
	\label{tab:models_performance}%
\end{table*}%

\begin{table}[htbp]
	\centering
	\scriptsize
	\caption{Class wise model performance}
	\begin{tabular}{lccccc}
		\toprule
		&       & Precision & Recall & f$_1$-score & Support \\
		\midrule
		\multicolumn{1}{c}{\multirow{4}[1]{*}{BNN}} & \multicolumn{1}{c}{0} & \multicolumn{1}{c}{0.61} & \multicolumn{1}{c}{0.47} & \multicolumn{1}{c}{0.53} & \multicolumn{1}{c}{419368} \\
		\multicolumn{1}{c}{} & \multicolumn{1}{c}{1} & \multicolumn{1}{c}{0.68} & \multicolumn{1}{c}{0.79} & \multicolumn{1}{c}{0.73} & \multicolumn{1}{c}{590635} \\
		\multicolumn{1}{c}{} & \multicolumn{1}{c}{macro avg} & \multicolumn{1}{c}{0.65} & \multicolumn{1}{c}{0.63} & \multicolumn{1}{c}{0.63} & \multicolumn{1}{c}{1010003} \\
		\multicolumn{1}{c}{} & \multicolumn{1}{c}{weighted avg} & \multicolumn{1}{c}{0.65} & \multicolumn{1}{c}{0.66} & \multicolumn{1}{c}{0.65} & \multicolumn{1}{c}{1010003} \\
		\midrule
		\multicolumn{1}{c}{\multirow{3}[0]{*}{DNN}} & \multicolumn{1}{c}{0} & \multicolumn{1}{c}{0.56} & \multicolumn{1}{c}{\textbf{0.74}} & \multicolumn{1}{c}{\textbf{0.64}} & \multicolumn{1}{c}{419368} \\
		\multicolumn{1}{c}{} & \multicolumn{1}{c}{1} & \multicolumn{1}{c}{\textbf{0.76}} & \multicolumn{1}{c}{0.59} & \multicolumn{1}{c}{0.67} & \multicolumn{1}{c}{590635} \\
		\multicolumn{1}{c}{} & \multicolumn{1}{c}{avg} & \multicolumn{1}{c}{0.68} & \multicolumn{1}{c}{0.65} & \multicolumn{1}{c}{0.66} & \multicolumn{1}{c}{1010003} \\
		\midrule
		\multicolumn{1}{c}{\multirow{4}[0]{*}{RF}} & \multicolumn{1}{c}{0} & \multicolumn{1}{c}{0.64} & \multicolumn{1}{c}{0.46} & \multicolumn{1}{c}{0.54} & \multicolumn{1}{c}{419368} \\
		\multicolumn{1}{c}{} & \multicolumn{1}{c}{1} & \multicolumn{1}{c}{0.68} & \multicolumn{1}{c}{\textbf{0.82}} & \multicolumn{1}{c}{0.74} & \multicolumn{1}{c}{590635} \\
		\multicolumn{1}{c}{} & \multicolumn{1}{c}{macro avg} & \multicolumn{1}{c}{0.66} & \multicolumn{1}{c}{0.64} & \multicolumn{1}{c}{0.64} & \multicolumn{1}{c}{1010003} \\
		\multicolumn{1}{c}{} & \multicolumn{1}{c}{weighted avg} & \multicolumn{1}{c}{0.67} & \multicolumn{1}{c}{0.67} & \multicolumn{1}{c}{0.66} & \multicolumn{1}{c}{1010003} \\
		\midrule
		\multicolumn{1}{c}{\multirow{4}[0]{*}{XGB}} & \multicolumn{1}{c}{0} & \multicolumn{1}{c}{\textbf{0.65}} & \multicolumn{1}{c}{0.53} & \multicolumn{1}{c}{0.58} & \multicolumn{1}{c}{419368} \\
		\multicolumn{1}{c}{} & \multicolumn{1}{c}{1} & \multicolumn{1}{c}{0.7} & \multicolumn{1}{c}{0.8} & \multicolumn{1}{c}{\textbf{0.75}} & \multicolumn{1}{c}{590635} \\
		\multicolumn{1}{c}{} & \multicolumn{1}{c}{macro avg} & \multicolumn{1}{c}{\textbf{0.68}} & \multicolumn{1}{c}{\textbf{0.66}} & \multicolumn{1}{c}{\textbf{0.67}} & \multicolumn{1}{c}{1010003} \\
		\multicolumn{1}{c}{} & \multicolumn{1}{c}{weighted avg} & \multicolumn{1}{c}{\textbf{0.68}} & \multicolumn{1}{c}{\textbf{0.69}} & \multicolumn{1}{c}{\textbf{0.68}} & \multicolumn{1}{c}{1010003} \\
		\midrule
		\multicolumn{1}{c}{\multirow{4}[0]{*}{DT}} & \multicolumn{1}{c}{0} & \multicolumn{1}{c}{0.6} & \multicolumn{1}{c}{0.59} & \multicolumn{1}{c}{0.59} & \multicolumn{1}{c}{419368} \\
		\multicolumn{1}{c}{} & \multicolumn{1}{c}{1} & \multicolumn{1}{c}{0.71} & \multicolumn{1}{c}{0.71} & \multicolumn{1}{c}{0.71} & \multicolumn{1}{c}{590635} \\
		\multicolumn{1}{c}{} & \multicolumn{1}{c}{macro avg} & \multicolumn{1}{c}{0.65} & \multicolumn{1}{c}{0.65} & \multicolumn{1}{c}{0.65} & \multicolumn{1}{c}{1010003} \\
		\multicolumn{1}{c}{} & \multicolumn{1}{c}{weighted avg} & \multicolumn{1}{c}{0.66} & \multicolumn{1}{c}{0.66} & \multicolumn{1}{c}{0.66} & \multicolumn{1}{c}{1010003} \\
		\midrule
		\multicolumn{1}{c}{\multirow{4}[0]{*}{LR}} & \multicolumn{1}{c}{0} & \multicolumn{1}{c}{0.58} & \multicolumn{1}{c}{0.53} & \multicolumn{1}{c}{0.56} & \multicolumn{1}{c}{419368} \\
		\multicolumn{1}{c}{} & \multicolumn{1}{c}{1} & \multicolumn{1}{c}{0.69} & \multicolumn{1}{c}{0.73} & \multicolumn{1}{c}{0.71} & \multicolumn{1}{c}{590635} \\
		\multicolumn{1}{c}{} & \multicolumn{1}{c}{macro avg} & \multicolumn{1}{c}{0.64} & \multicolumn{1}{c}{0.63} & \multicolumn{1}{c}{0.63} & \multicolumn{1}{c}{1010003} \\
		\multicolumn{1}{c}{} & \multicolumn{1}{c}{weighted avg} & \multicolumn{1}{c}{0.64} & \multicolumn{1}{c}{0.65} & \multicolumn{1}{c}{0.65} & \multicolumn{1}{c}{1010003} \\
		\midrule
		\multicolumn{1}{c}{\multirow{4}[1]{*}{MNB}} & \multicolumn{1}{c}{0} & \multicolumn{1}{c}{0.45} & \multicolumn{1}{c}{0.53} & \multicolumn{1}{c}{0.49} & \multicolumn{1}{c}{419368} \\
		\multicolumn{1}{c}{} & \multicolumn{1}{c}{1} & \multicolumn{1}{c}{0.62} & \multicolumn{1}{c}{0.54} & \multicolumn{1}{c}{0.58} & \multicolumn{1}{c}{590635} \\
		\multicolumn{1}{c}{} & \multicolumn{1}{c}{macro avg} & \multicolumn{1}{c}{0.54} & \multicolumn{1}{c}{0.54} & \multicolumn{1}{c}{0.53} & \multicolumn{1}{c}{1010003} \\
		\multicolumn{1}{c}{} & \multicolumn{1}{c}{weighted avg} & \multicolumn{1}{c}{0.55} & \multicolumn{1}{c}{0.54} & \multicolumn{1}{c}{0.54} & \multicolumn{1}{c}{1010003} \\
		\bottomrule
	\end{tabular}%
	\label{tab:class_wise_models_performance}%
\end{table}%

\subsection{XGBoost Feature Identification}

Every customer is unique and has varying financial situations and needs. Payment plans and other assistance methods need to be considered depending on the customers individual situation given the fact that no single solution suits all situations. XGBoost is a decision tree-based model that provides comprehensibility to the decision making process. It highlights the characteristics and circumstances of customers that distinguish a customer, who is able to make the payment by the due date, with the one who is not able to meet the payment date. The model details the characteristics (or features) that describe a specific customer to be identified as prone to propensity-to-pay. Focusing on these features while proposing payment assistance programs can prevent premature outsourcing and unnecessary high vendor fees that lack return on investment. Some of these features identified by model are \emph{a specific customer age range, month and year of a bill, weekly median household income in the living area, bill duration, average household size in the living area} and \emph{remoteness of the living area}. 

\subsection{BNN Uncertainty Estimation}
\begin{figure*}[htb!]
	\begin{subfigure}{\textwidth}
		\includegraphics[scale=0.40]{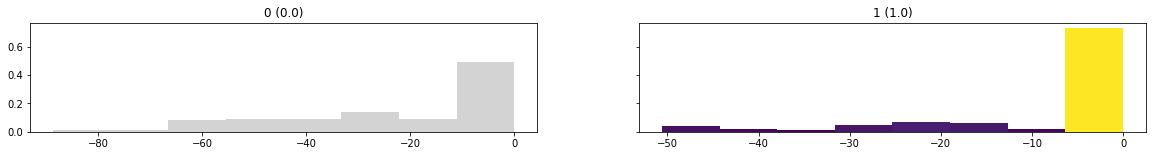}
		\caption{Prediction for customer 1: Correct and Low Uncertainty}
		\label{fig:clu}
	\end{subfigure} \hspace{0.2\textwidth}
	\begin{subfigure}{\textwidth}
		\includegraphics[scale=0.40]{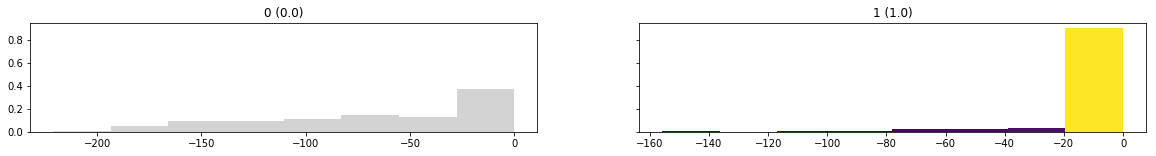}
		\caption{Prediction for customer 2: Correct and High Uncertainty}
		\label{fig:chu}
	\end{subfigure}
	\begin{subfigure}{\textwidth}
		\includegraphics[scale=0.40]{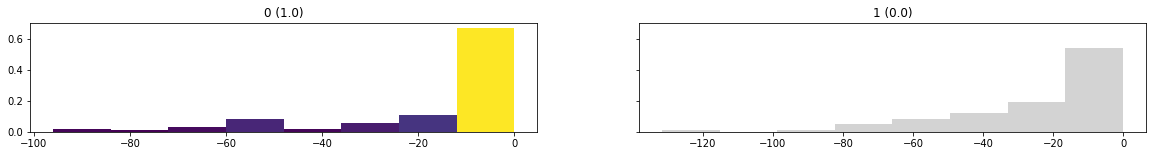}
		\caption{Prediction for customer 3: Incorrect and High Uncertainty}
		\label{fig:ichu}
	\end{subfigure} \hspace{0.2\textwidth}
	\begin{subfigure}{\textwidth}
		\includegraphics[scale=0.40]{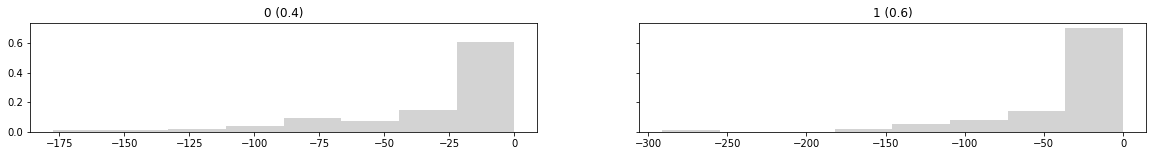}
		\caption{Prediction for customer 4: High Uncertainty and Undecided}
		\label{fig:huu}
	\end{subfigure}
	\caption{Four Example Test Instances Classified by BNN. Horizontal axis shows log-probability and vertical axis shows frequency of the corresponding probability.}
	\label{fig:bnn_class_dist}
\end{figure*}

For each test instance (i.e. a (new) bill unforeseen to the model), BNN shows a histogram of log-probabilities for each of two classes. Some examples are given in Figure \ref{fig:bnn_class_dist}. In this figure, horizontal axis shows log-probability and vertical axis shows frequency of the corresponding probability. Log-probability close to 0 means the probability is close to 1 $(exp(0) = 1)$. We take median of the probability distribution as the probability for each class. If this probability for class $C_k$ is greater than a threshold, we map this probability to the class $C_k$. When the mapped class $C_k$ is same as the real class, it is \emph{correct}. Yellow coloured bar indicates that the model has mapped the probability to the class, while grey coloured bars indicate that the model is undecided. 

Figure \ref{fig:bnn_class_dist} shows the histograms by plotting the distribution of log-probabilities for two classes. The wider histograms in Figure \ref{fig:ichu} show that the network has a high uncertainty for both classes 0 and 1 when the prediction is incorrect or the network is undecided as shown in Figure \ref{fig:huu}. Whereas in the case of the correct prediction, the log-probability distribution for the correct class is narrow while for the other class it is wide (as shown in Figures \ref{fig:clu} and \ref{fig:chu}), which means the network is more certain of the correct class.

To see how the prediction of BNN is superior to other models, consider Figures \ref{fig:clu} and \ref{fig:chu}. Here the model predicts with the highest probability  (i.e., a value of 1 in vertical axis - yellow bar) that both customers will pay their bill on time. In traditional NN models, we have only this information available and we make decisions based on this information only. However, BNN tells us that the model is more certain about the prediction of the first customer (\ref{fig:clu}) because its distribution is narrower (-50 to 0) in comparison to that of the second customer (-160 to 0) (\ref{fig:chu}).

When the uncertainty of a prediction is high, it often results in a wrong prediction even though the model considers the probability is high. For example, consider Figures \ref{fig:chu} and \ref{fig:huu}, the outcomes of both customers are predicted with the same probability 1 (100\%). However, one of them is wrongly predicted (Figure \ref{fig:huu}) and it can only be confirmed by checking the high uncertainty attached to the prediction. More importantly, BNN can utilise the probability distribution to decide when not to predict. For example, BNN decided not to predict the class of the forth customer in Figure \ref{fig:huu}.

\subsection{Deployment: Model Implication}
Due to the essential nature of businesses along with the social and community impact of customers struggling or missing bill payments, it can prove beneficial to both the individual customers and the wider community to take proactive action based on a good understanding of a customer's propensity-to-pay and thus to reduce negative efforts. Machine learning models can be deployed at scale and the outcomes can be obtained for each customer (including new customers as the model generalises the findings from the historical data to be applied to new data). 

Predicting propensity-to-pay is often not a simple process. Access to a large number of features and data is essential for models to provide accurate predictions. In practice, many important features are either not recorded or protected from access because of a desire to protect the customers privacy and prevent unwarranted intrusion. Our experiments with decision tree models inform the client to focus on useful features for implementing this proposed solution and make the data collection process easier and viable. When the right number of features and data are available, BNN can provide prediction with higher accuracy and confidence. 

Since communication with, and providing education to customers adds additional costs, utilising the uncertainty estimation by the deployed BNN model is worth consideration. BNN can separate customers into three categories: (a) will definitely pay, (b) will definitely not pay and (c) not sure about the payment. This separation allows action to be focused in the most efficient way through the use of cost-benefit estimations before acting on the ``not sure" group. 

Separately to the energy bill scenario we discuss above; some financial institutions have started to predict customers' financial behaviour using machine learning algorithms to create profiles for \emph{ideal} customers and customising operations to nurture customers with higher lifetime values \cite{moradi2019dynamic}. Adapting BNN could provide them better confidence in constructed profiles. 

Though there are some challenges in implementing propensity-to-pay models in practice, the benefits these models provide are measurable across cost reduction, revenue and customer satisfaction levels.

\section{Conclusion}
\label{sec:conclusion}
This study investigates machine learning models' ability to consider different contexts and
estimate the uncertainty in the prediction. Seven models from four family of machine learning algorithms are investigated for their novel utilisation. Especially, this paper introduces a novel concept of utilising a Baysian Neural Network to a binary classification problem and applies this to a real-world problem of predicting propensity-to-pay energy bills. 
We used the limited number of variables that the organisation can source observing the data privacy and data collection restrictions. Using the currently available features in variety of settings and analysing their corresponding prediction accuracy, we identified the predictive power of certain features that can be utilised by the organisation easily in deployment.

We observed that DNN is high in precision, RF is high in recall and XGB is high in accuracy; and when an average over all the measures (i.e. Accuracy, Precision, Recall, F1 score, Cohen Kappa and Area Under Curve) is taken, BNN performs better than DNN. This means an organisation can consider one or more models from BNN, DNN, XGB and RF depending on their objectives when utilising the prediction. For example, besides high performance, BNN can provide uncertainty estimation; outcomes from RF and XGB are easily explainable and can identify and visualise the critical and influencing features. 

The goal of prediction is to help organisations understand customers in hardship and take proactive action to mitigate negative effects leading to improved customer satisfaction and better outcomes. This knowledge of customers propensity-to-pay based on the prediction of  machine learning models can be used to assist vulnerable customers and empower them with options that suit their situation. The prediction algorithms combines public (ABS data), customer (account information), and proprietary data (bill information) to estimate the probability that a customer may fail to pay their bill by the due date. In general this kind of prediction can enable organisations such as periodic service providers, payment solutions, lending institutions and health care systems to calculate the financial risk of service transactions and make new financing options and educational information available to customers. 

\section*{Funding}
Funding was provided through the Energy Queensland Innovation Hub to explore these ideas on behalf of its Ergon Energy Retail subsidiary.

\bibliographystyle{apacite}
\bibliography{References}


\vspace{10mm}
\appendixname{ A: Description of Evaluation Measures}

\begin{itemize}
    \item True Positive (TP): True positives are instances classified as positive by the model that actually are positive. 
    \item True Negative (TN): True negatives are instances the model classifies as negative that actually are negative. 
    \item False Positive (FP): False positives are instances identified by model as positive that actually are negative.
    \item False Negative (FN): False negatives are instances the model classifies as negative that actually are positive.  
    \item Accuracy: It is the percentage of correctly classified instances, and it is calculated as $\frac{TP + TN}{TP + TN + FP + FN}$.
    \item Precision: It calculates a model's ability to return only relevant instances. It is calculated as $\frac{TP}{TP + FP}$. 
    \item Recall: It calculates a model's ability to identify all relevant instances. It is calculated as 
$\frac{TP}{TP + FN}$. 
    \item $F_1$ Score: A single metric that combines recall and precision using the harmonic mean. $F_1$ Score is calculated as $2 \times \frac{precision}{precision + recall}$.
    \item Cohen Kappa (CK):  Cohen's kappa score is used to measure inter-rater and itra-rater reliability for categorical items \cite{mchugh2012interrater}. It is calculated as $\frac{OA-AC}{1-AC}$, where where $OA$ is the relative observed agreement between predicted labels and actual labels and $AC$ is the probability of agreement by chance. 
    \item Area Under Curve (AUC): Area under the Receiver operating characteristic (ROC) curve is called Area Under the Curve (AUC). ROC plots the true positive rate  versus the false positive rate as a function of the model’s threshold for classifying a positive. AUC calculates the overall performance of a classification model.
\end{itemize}

\end{document}